\documentclass{article}

\PassOptionsToPackage{numbers, compress}{natbib}
\usepackage[preprint]{arxiv}

\usepackage[utf8]{inputenc} % allow utf-8 input
\usepackage[T1]{fontenc} % use 8-bit T1 fonts
\usepackage{hyperref} % hyperlinks
\usepackage{url} % simple URL typesetting
\usepackage{booktabs} % professional-quality tables
\usepackage{amsfonts} % blackboard math symbols
\usepackage{nicefrac} % compact symbols for 1/2, etc.
\usepackage{microtype} % microtypography
\usepackage{xcolor} % colors
\usepackage{enumitem}
\usepackage{xspace}
\usepackage{etoolbox}
\usepackage{titletoc} % Load the package
\usepackage{multicol}
\usepackage{multirow}
\usepackage{amssymb}
\usepackage{wrapfig}
\usepackage{pifont}
\usepackage{bbding}
\usepackage{color, colortbl}
\usepackage{subcaption}
\usepackage{url}
\usepackage{graphicx}
\usepackage{listings}
\usepackage{xcolor}
\usepackage{colortbl}
\usepackage{rotating}
\usepackage{hyperref}
\usepackage{colortbl}
\usepackage{booktabs}
\usepackage{tabularx}
\usepackage{makecell}
\usepackage{bbding}
\usepackage{listings}
\usepackage{amsthm}
\usepackage{fdsymbol}
\usepackage{amsmath,amsfonts,bm}

\def\eqref#1{equation~\ref{#1}}
\def\1{\bm{1}}

\def\rmH{{\mathbf{H}}}

\def\rmK{{\mathbf{K}}}

\def\rmM{{\mathbf{M}}}

\def\rmO{{\mathbf{O}}}

\def\rmS{{\mathbf{S}}}

\def\rmV{{\mathbf{V}}}

\def\rmX{{\mathbf{X}}}

\def\rmZ{{\mathbf{Z}}}

\DeclareMathAlphabet{\mathsfit}{\encodingdefault}{\sfdefault}{m}{sl}
\SetMathAlphabet{\mathsfit}{bold}{\encodingdefault}{\sfdefault}{bx}{n}

\def\gA{{\mathcal{A}}}
\def\gB{{\mathcal{B}}}

\def\gE{{\mathcal{E}}}

\def\gG{{\mathcal{G}}}
\def\gH{{\mathcal{H}}}

\def\gL{{\mathcal{L}}}

\def\gN{{\mathcal{N}}}

\def\gR{{\mathcal{R}}}

\def\gT{{\mathcal{T}}}

\def\gV{{\mathcal{V}}}

\def\gX{{\mathcal{X}}}

\def\gZ{{\mathcal{Z}}}

\def\sR{{\mathbb{R}}}

\newcommand{\KL}{D_{\mathrm{KL}}}

\DeclareMathOperator{\Tr}{Tr}

\usepackage[most]{tcolorbox}
\usepackage{tikz}
\usetikzlibrary{tikzmark}
\makeatletter
\newcommand*{\myfontsize}{%
\@setfontsize\myfontsize{6.7}{8}%
}
\makeatother

\definecolor{cadmiumgreen}{rgb}{0.0, 0.42, 0.24}

\definecolor{myred}{rgb}{0.7, 0.3, 0.0}
\definecolor{myblue}{rgb}{0.2, 0.3, 0.6}

\newcommand{\modelname}{BHyGNN+\xspace}
\definecolor{green(pigment)}{rgb}{0.0, 0.65, 0.31}
\definecolor{darksalmon}{rgb}{0.91, 0.59, 0.48}

\definecolor{Gray}{gray}{0.95}
\definecolor{royalblue(web)}{rgb}{0.25, 0.41, 0.88}
\definecolor{codegreen}{rgb}{0,0.6,0}
\definecolor{codegray}{rgb}{0.5,0.5,0.5}
\definecolor{codepurple}{rgb}{0.58,0,0.82}

\theoremstyle{definition}
\newtheorem{definition}{Definition}[section]

\lstdefinestyle{mystyle}{%
language=Python, commentstyle=
\color{codegreen}
, keywordstyle=
\color{magenta}
, numberstyle=\tiny
\color{codegray}
, stringstyle=
\color{codepurple}
,
basicstyle = \ttfamily \lst@ifdisplaystyle\small\fi, breakatwhitespace=false, breaklines=true,
captionpos=b, keepspaces=true, numbers=left, numbersep=5pt, xleftmargin=12pt, showspaces=false,
showstringspaces=false, showtabs=false, tabsize=2,
moredelim=[is][\bfseries]{<highlight>}{</highlight>}, %
postbreak=\raisebox{0ex}[0ex][0ex]{\ensuremath{\color{black}
\lst@ifdisplaystyle\hookrightarrow\fi\space}} %
}
\title{\modelname: Unsupervised Representation Learning\\
for Heterophilic Hypergraphs}

\author{
    Tianyi Ma$^{\spadesuit}$,
    Yiyue Qian$^{\clubsuit}$\thanks{The work is not related to the position at Amazon.},
    Zehong Wang$^{\spadesuit}$,
    Zheyuan Zhang$^{\spadesuit}$, 
    Chuxu Zhang$^{\vardiamondsuit}$,
    Yanfang Ye$^{\spadesuit}$\thanks{Corresponding author.}\\ \\
    $^\spadesuit$\textit{University of Notre Dame,}
    $ ^\vardiamondsuit$ \textit{University of Connecticut,}
    $^\clubsuit$\textit{Amazon GenAI},
    \\ \\
    {tma2@nd.edu, yyqian5@gmail.com, \{zwang43, zzhang42\}@nd.edu} \\
    {chuxu.zhang@uconn.edu, yye7@nd.edu.}
}

\begin{document}
  \maketitle
  \begin{abstract}
    Hypergraph Neural Networks (HyGNNs) have demonstrated remarkable success in modeling higher-order relationships among entities. 
However, their performance often degrades on heterophilic hypergraphs, where nodes connected by the same hyperedge tend to have dissimilar semantic representations or belong to different classes. 
While several HyGNNs, including our prior work BHyGNN, have been proposed to address heterophily, their reliance on labeled data significantly limits their applicability in real-world scenarios where annotations are scarce or costly.
To overcome this limitation, we introduce \textbf{\modelname}, a self-supervised learning framework that extends BHyGNN for representation learning on heterophilic hypergraphs without requiring ground-truth labels. 
The core idea of \modelname is \textbf{hypergraph duality}, a structural transformation where the roles of nodes and hyperedges are interchanged.
By contrasting augmented views of a hypergraph against its dual using cosine similarity, our framework captures essential structural patterns in a fully unsupervised manner. 
Notably, this duality-based formulation eliminates the need for negative samples, a common requirement in existing hypergraph contrastive learning methods that is often difficult to satisfy in practice.
Extensive experiments on eleven benchmark datasets demonstrate that \modelname consistently outperforms state-of-the-art supervised and self-supervised baselines on both heterophilic and homophilic hypergraphs. 
Our results validate the effectiveness of leveraging hypergraph duality for self-supervised learning and establish a new paradigm for representation learning on challenging, unlabeled hypergraphs.

  \end{abstract}

\section{Introduction}

Hypergraphs provide a powerful formalism for modeling complex, higher-order relationships in data, where a single interaction can involve more than two entities. 
This expressive capability has driven the development of numerous Hypergraph Neural Networks (HyGNNs) designed to learn effective representations from such structured data~\cite{HGNN, HyperGCN, Allset}. 
However, a large portion of existing HyGNNs is built upon the assumption of \textit{homophily}, i.e., the principle that nodes connected within a hyperedge are likely to share similar semantic representations or belong to the same class~\cite{ma2021homophily, platonov2023critical}. 
These models typically employ message-passing neural networks~\cite{gilmer2020message, wang2025graph} that propagate and aggregate information from neighboring nodes, a mechanism that naturally reinforces similarity and performs well in homophilic settings.

However, real-world data frequently exhibit \textit{heterophily}, where connected nodes are dissimilar or belong to different classes~\cite{zhu2020beyond}. This phenomenon is particularly prevalent in hypergraphs.
For example, in a co-purchase dataset, a single shopping cart (hyperedge) often contains products spanning diverse categories (nodes)~\cite{amburg2020clustering}. 
In such heterophilic settings, standard message-passing mechanisms can be counterproductive: by aggregating features from dissimilar neighbors, they risk blurring the distinctions between nodes of different classes and degrading downstream performance. 
Despite its practical importance, learning effective representations from heterophilic hypergraphs remains a critical yet underexplored challenge~\cite{wang2022equivariant, HyperGCL}.

To address this gap, our prior conference work~\cite{ma2025broadcast} introduced the \textbf{B}roadcast \textbf{Hy}per\textbf{G}raph \textbf{N}eural \textbf{N}etwork (BHyGNN), a supervised framework tailored for heterophilic hypergraphs. 
BHyGNN introduces an adaptive propagation mechanism in which nodes learn to either \textit{broadcast} their features to a hyperedge or \textit{receive} information from it. 
This selective propagation is enabled by a \textbf{V}ariational \textbf{B}roadcast \textbf{A}utoencoder Network (VBA-Net), which works in conjunction with an incorporation transformer to learn discriminative node representations. 
The entire model is trained end-to-end, relying on downstream class labels to guide the learning of both the propagation actions and the node representations.

While effective, the supervised nature of BHyGNN presents a significant limitation: 
\textbf{it requires labeled data, which can be scarce or expensive to obtain in many real-world applications} \cite{qian2021distilling, wang2025can, fan2018automatic, ma-etal-2025-llm, zhang2025mopi}.
This limitation motivates the primary contribution of this journal extension: \textbf{\modelname}, a self-supervised framework that learns informative representations for heterophilic hypergraphs without relying on ground-truth labels.
\modelname extends BHyGNN into a contrastive self-supervised learning (SSL) paradigm by leveraging the concept of \textbf{hypergraph duality}~\cite{bretto2013hypergraph, aksoy2020hypernetwork}.
In a dual hypergraph, the roles of nodes and hyperedges are interchanged: each hyperedge in the original hypergraph becomes a node in the dual, and each original node becomes a hyperedge connecting the corresponding dual nodes.
This transformation offers a complementary structural perspective, revealing relationships between hyperedges that are not apparent from the node-centric view alone~\cite{berge1985graphs, bretto2013applications}. 
Crucially, by framing self-supervised learning as a contrastive task between a hypergraph and its dual, \modelname eliminates the need for negative samples, i.e., a common requirement in hypergraph contrastive learning that is often difficult to fulfill in practice. 
In summary, the main contributions of this work are as follows:
\begin{itemize}[leftmargin=*]
    \setlength{\itemsep}{0pt}
    \item \textbf{Problem:} We address the challenging problem of representation learning on heterophilic hypergraphs in a practical, unsupervised setting where labels are unavailable during training.
    \item \textbf{Novelty:} 
    We propose \modelname, a novel self-supervised learning framework that leverages \textbf{hypergraph duality} to learn representations for heterophilic hypergraphs. 
    To the best of our knowledge, this is the first work to exploit hypergraph duality for self-supervised learning tailored to heterophilic settings. 
    Furthermore, our duality-based contrastive formulation naturally eliminates the need for negative samples, circumventing a key challenge in existing hypergraph contrastive learning methods.
    \item \textbf{Extensive Evaluation:} We conduct comprehensive experiments on eleven benchmark datasets, including both heterophilic and homophilic hypergraphs as well as a synthetic heterophilic dataset. 
    The results demonstrate that \modelname consistently achieves state-of-the-art performance, outperforming a wide range of supervised and self-supervised baselines.
\end{itemize}

\section{Background}
\subsection{Preliminary}

\begin{definition}\textbf{Hypergraph.} Given a hypergraph $\mathcal{H}=\left (\mathcal{V}, \mathcal{E}, \mathcal{X}\right )$, $\mathcal{V} = \left\{v_1, \dots, v_N\right\}$ is the set of nodes 
of size $N$, $\mathcal{E} = \left\{e_1, \dots, e_M\right\}$ is the set of hyperedges of size $M$, and $\mathcal{X}$ is the feature set for nodes and hyperedges, i.e., $\gX = \gX_\gV \cup \gX_\gE$. 
A hypergraph is represented by an incidence matrix $\mathbf{M} \in \mathbb{R}^{N\times M}$, 
where $\mathbf{M}_{i,j} = 1$ if $ v_i\in e_j$; otherwise, $\mathbf{M}_{i,j} = 0$. 
% Here node $v \in \mathcal{V}$ and hyperedge $e \in \mathcal{E}$. 
Additionally, $d(v_i) = \sum_{e_j\in\mathcal{E}} \mathbf{M}_{i,j}$ and  $d(e_j) = \sum_{v_i\in\mathcal{V}} \mathbf{M}_{i,j}$ denote the degrees of a node and a hyperedge, respectively. 
\end{definition}

\begin{definition}\textbf{Hypergraph Homophily.} 
Given a hypergraph $\mathcal{H}=(\mathcal{V}, \mathcal{E}, \mathcal{X})$ with a set of node classes $\mathcal{C} = \{c_1, \dots, c_{|\mathcal{C}|}\}$, the homophily score of the hypergraph $\gH$ is measured from node and hyperedge aspects. The homophily score for node $v$ is computed as 
    $h(v) = |\{u: u\in\gN_v\land y_v = y_u\}| /|\gN_v|,$
where $y_v$ is the label of node $v$, and $\gN_v$ is the neighbors of node $v$. 
The homophily score of a hyperedge $e$ is the maximum ratio of nodes in hyperedge $e$ that have the same class label:
    $h(e) = \max_{c\in\mathcal{C}} |\{v:v\in e \land y_v = c \}| /  |e| .$
We refer to the class that emits the maximum ratio as the majority class of hyperedge $e$, and the rest of the classes are the minority classes of hyperedge $e$. 
\end{definition}
\begin{definition}\label{def: se-based-hygnn}
    \textbf{SE-based Hypergraph Neural Networks.} 
Given a hypergraph $\mathcal{H}=(\mathcal{V}, \mathcal{E}, \mathcal{X})$, the star expansion (SE)-based HyGNNs first convert the hypergraph $\gH$ into a bipartite graph $\mathcal{G}=(\mathcal{V}\cup \mathcal{E}, \mathcal{R}, \mathcal{X}')$, where $\mathcal{R} = \{(v, e)| v\in e, v\in \mathcal{V}, e\in \mathcal{E}\}$ is the set of edges connecting nodes and hyperedges, and $\mathcal{X}'$ is the attribute feature set for nodes and hyperedges. 
Then, HyGNNs leverage a message-passing function $f_*(\cdot)$ to learn hyperedge and node representations iteratively:
\begin{equation}
    \rmZ_\gE = f_{\gV\rightarrow \gE}(\rmX_\gV, \rmX_\gE, \rmM_\gR), \,
    \rmZ_\gV = f_{\gE\rightarrow \gV}(\rmZ_\gE, \rmX_\gV, \rmM_\gR),
\end{equation}
Here, $\rmX_\gV$ and $\rmX_\gE$ are the feature matrices for nodes and hyperedges, respectively, and $\rmM_\gR$ is the incidence matrix of the bipartite graph $\gG$. Note that $\rmZ_\gE$ and $\rmZ_\gV$ are the learned representations for hyperedges and nodes, respectively.
\end{definition}
\begin{definition} \label{def: dual-hypergraph}
    \textbf{Dual Hypergraph.}
    Given a hypergraph $\gH = (\gV, \gE, \gX)$, its dual hypergraph is defined as $\gH^* = \Tr(\gH) = (\gE, \gV, \gX^*)$, where each hyperedge in $\gH$ corresponds to a node in $\gH^*$, and each node in $\gH$ corresponds to a hyperedge in $\gH^*$. 
    The node feature matrix $\gX_\gV^* \in \gX^*$ is derived from the hyperedge feature matrix $\gX_\gE \in \gX$, and vice versa.
\end{definition}

\noindent\textbf{Problem Definition.}
\textit{
\textbf{Heterophilic Hypergraph Contrastive Learning.}
Given a heterophilic hypergraph $\mathcal{H} = (\mathcal{V}, \mathcal{E}, \mathcal{X})$,
the objective is to build a hypergraph neural network model $f: \mathcal{V}\rightarrow\mathbb{R}^d$ to project nodes into $d$-dimensional embeddings that are representative for downstream tasks.
}

\subsection{Related Work}

\textbf{Hypergraph Neural Networks.} 
Unlike graphs where each edge connects two nodes~\cite{wang2025can, wang2025generative, GCN, ma2025autodata, qian2021adapting,wen2024gcvr, wang2025beyond}, hypergraphs allow hyperedges to connect an arbitrary number of nodes, enabling the representation of higher-order relationships~\cite{ma2025adaptive,wang2026molecular, qian2025enhancingecommerce}. 
To utilize the advantages of hypergraphs, various Hypergraph Neural Networks (HyGNNs)~\cite{Allset,ma2021homophily,platonov2023critical,lim2021large, HCHA, HNHN} have been developed, aiming to model the complex relationships in hypergraphs effectively.
Early HyGNNs~\cite{HyperGCN, HGNN} attempted to apply GNNs to hypergraphs by first transforming hypergraphs into graphs and then utilizing GNNs to learn node representations. 
Despite the simplicity of these early studies, they usually lead to undesired losses in learning performance~\cite{Allset,pmlr-v89-chien19a}.
On the other hand, recent works~\cite{Allset,wang2022equivariant} view hypergraphs as bipartite graphs and design multiset propagation rules that generate the hyperedge embeddings from node attribute features and propagate hyperedge embeddings back to nodes as final node embeddings.
Inspired by these advances, we propose an adaptive hypergraph propagation mechanism that selectively propagates information between nodes and hyperedges. 

\textbf{Heterophilic Graph Representation Learning.}
In recent years, various graph representation learning models have been proposed to address the heterophily settings~\cite{zhu2021graph,chien2020adaptive,wang2022powerful, wang2025towards}. 
For instance, Geom-GCN~\cite{pei2020geom} pre-computes unsupervised node embeddings and leverages geometric relationships in the embedding space to establish a bi-level aggregation process.
Additionally, a more recent study~\cite{bodnar2022neural} introduces neural sheaf diffusion models that learn cellular sheaves to address the heterophily issue in graphs. 
Despite this progress on graphs \cite { zhao2021multi, zhang2025agentrouter, ju2022adaptive,ju2022grape, qian2022co, zhao2023self, wang2024gft}, representation learning on heterophilic hypergraphs remains underexplored, motivating our work. 

\textbf{Hypergraph Contrastive Learning.}
To eliminate the dependency on label information, recent studies~\cite{HyperGCL, ma2023hypergraph, kim2024hypeboy, lee2023m} have explored hypergraph contrastive learning (HyGCL) that integrates contrastive learning with HyGNNs.
For example, TriCL~\cite{lee2023m} leverages tri-directional contrastive learning to capture node, group, and membership-level information within hypergraphs simultaneously. 
While these approaches have demonstrated improvements under scarce labels, they depend heavily on negative samples, whose high-quality construction is challenging, particularly in large hypergraphs. 
To address this limitation, we introduce a negative-sample-free HyGCL method that relies only on positive contrastive pairs for self-supervised learning.
\section{Methodology}
\begin{figure}
    \centering
    \includegraphics[width=\linewidth]{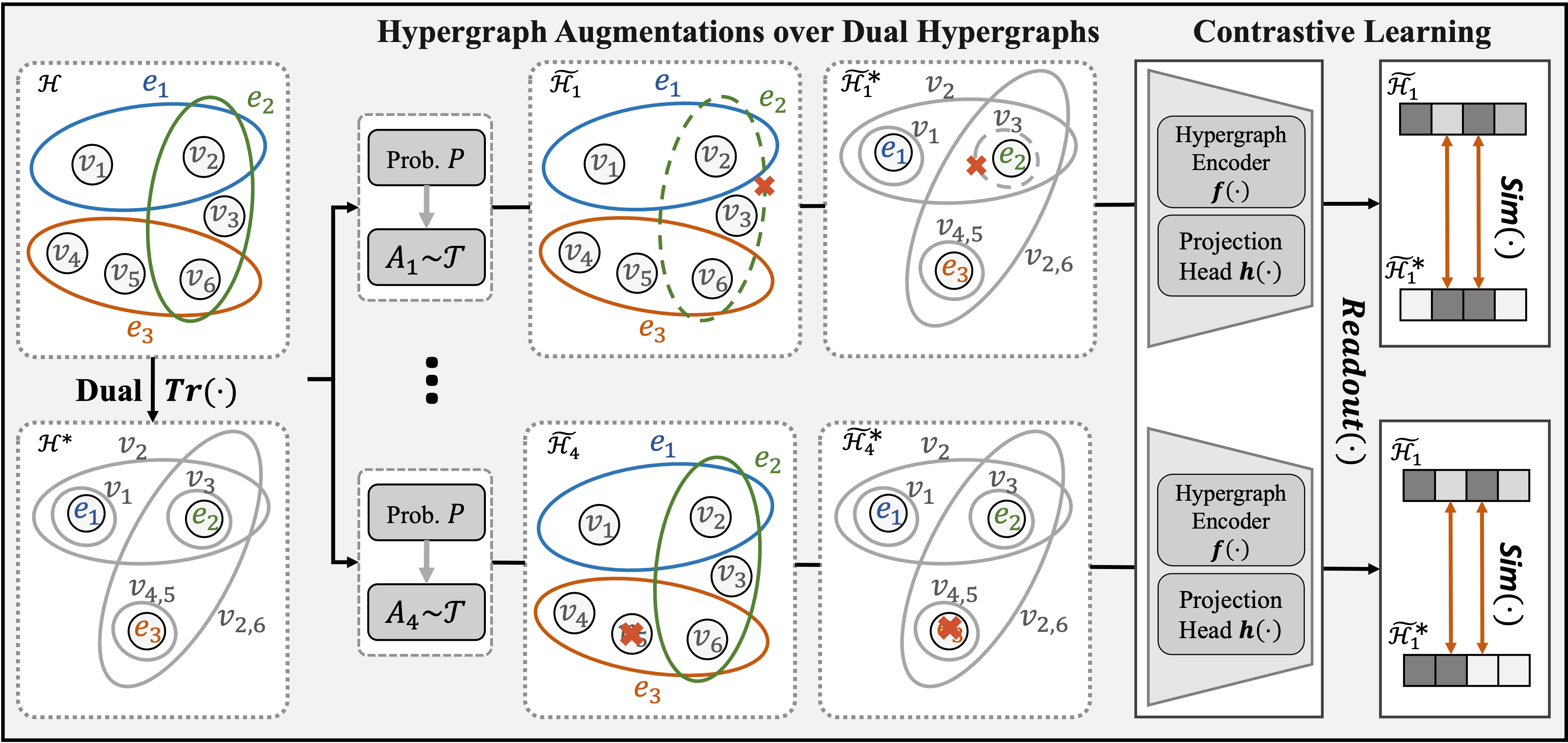}
    \caption{The overall framework of \modelname. Given a hypergraph $\gH$, we first obtain its dual $\gH^* = \Tr(\gH)$. For each augmentation method $A_i\in\gT$, we generate the augmented hypergraph pair $\{\widetilde{H_i}, \widetilde{H_i^*}\}$. Each augmented hypergraph is then fed into the hypergraph encoder $f(\cdot)$, a projection head $h(\cdot)$, and a readout function to obtain hypergraph-level representations. We use cosine similarity, i.e., $\text{sim}(\cdot)$, as the objective function to optimize the encoder.
    }
    \label{fig: framework}
\end{figure}
In this section, we present the details of BHyGNN and \modelname for heterophilic hypergraphs representation learning.
BHyGNN is introduced in our conference paper~\cite{ma2025broadcast}, while \modelname is the main contribution of this journal extension.
Note that our hypergraph convolutions follow the SE-based HyGNNs defined in Definition~\ref{def: se-based-hygnn}; that is, within one hypergraph convolution, we first update hyperedge representations $\rmZ_\gE'$, and then learn node representations $\rmZ_\gV'$ based on the updated hyperedge representations $\rmZ_\gE'$ and the original node representations $\rmZ_\gV$.

\subsection{BHyGNN for Supervised Learning}
We first discuss BHyGNN with the hypergraph convolution BHyGNNConv, which is proposed in our conference version paper~\cite{ma2025broadcast}. 
Since this is not the main contribution of this journal extension, we only present the mathematical formulation of BHyGNN and refer readers to our conference paper~\cite{ma2025broadcast} for additional details.
\subsubsection{Broadcast Hypergraph Convolution}
Given a hypergraph $\gH = (\gV, \gE,\gX)$, one BHyGNNConv layer is defined as:
\begin{align} \label{eq: conv}
    \rmZ_\gV',\rmZ_\gE' = \text{BHyGNNConv}(\gH, \rmZ_\gV, \rmZ_\gE), \\
  \rmZ_e' = \textbf{MH}(\rmS_e, \theta_\gE), \rmZ_v' = \textbf{MH}(\rmS_v, \theta_\gV),
\end{align}
where $\theta_* \in \sR^{1\times hd'}$ denotes the model parameters for multi-head attention \textbf{MH}$(\cdot)$, which we discuss in Equation~\ref{eq: mh}, and $\rmS_e\in \sR^{{c_e}\times d}$ and $\rmS_v\in \sR^{{c_v}\times d}$ are the stacked incorporation embeddings for $\gZ_e$ and $\gZ_v$, respectively. 
Mathematically, we have:
\begin{align}
    \gZ_e^{\scriptscriptstyle{(\textbf{B})}} = \{\rmZ_v || \rmZ_e:  a_{v,e}^{\scriptscriptstyle{\left(\textbf{B}\right )}} \in \gA_e^{\scriptscriptstyle{\left(\textbf{B}\right)}} \},  c_e = |\gZ_e^{\scriptscriptstyle{(\textbf{B})}}|, \\ 
    \gZ_{v}^{\scriptscriptstyle{(\textbf{R})}} = \{\rmZ_e'||\rmZ_v:  a_{v,e}^{\scriptscriptstyle{(\textbf{R})}}
    \in \gA_v^{\scriptscriptstyle{\left(\textbf{R}\right)}}
    \}, c_v = |\gZ_{v}^{\scriptscriptstyle{(\textbf{R})}}|.
\end{align}
Here,  \textbf{B} and \textbf{R} denote the broadcast and receive actions for propagation in $f_{\gV\rightarrow\gE}$ and $f_{\gE\rightarrow\gV}$, respectively.
 $\gA$ denotes a set of propagation actions, i.e., $\gA = \{a_{v,e} : v \in e\}$ and is learned via a \textbf{V}ariational \textbf{B}roadcast \textbf{A}utoencoder \textbf{NET}work (VBA-Net), i.e., $g(\cdot)$\footnote{For simplicity, we drop the superscripts for action set $\gA$, propagation actions $a_{v,e}$ and VBA-Net $g(\cdot)$.}, which is formulated as:
\begin{align}
    a_{v, e} &\in \gA = g(\gH, \rmZ_\gV, \rmZ_\gE), \\
    a_{v,e} &= \text{Gumbel-Softmax}\left(P(\{v,e\}| \rmH^{\gV}, \rmH^{\gE})\right), \\
    &= \text{Gumbel-Softmax}\left(\text{Sigmoid}(\rmH_v^{(\gV)T}\rmH^{(\gE)}_e)\right),
\end{align}
where a pair \{$\rmH^{(\gV)}, \rmH^{(\gE)}$\} denotes the latent representation of the hypergraph $\gH$. 
With the variational posterior $q(\cdot)$, parameterized by MLPs, and prior $p_*(\cdot) \sim \gN(0, I)$, each latent representation $\rmH^*$ is computed as:
\begin{align}
    \rmH^*&= \rho_* \odot \sigma_* + \mu_*, \,\,
    \rho_* \sim \gN(0, I), \\
    \rmH^*&\sim q(\rmH^*|\gH) = \gN(\mu_*, \sigma_*^2). 
\end{align}
In Equation~\ref{eq: conv}, $\textbf{MH}(\cdot)$ is a multi-head attention parameterized by $\theta\in\sR^{1\times hd'}$, which is formulated as follows:
\begin{align} \label{eq: mh}
    \textbf{MH}(\rmS, \theta) = ||_{i=1}^h\rmO^{(i)}, \rmO^{(i)} = \omega(\theta^{(i)}(\rmK^{(i)})^{\top})\rmV,
\end{align}
where $h$ denotes the number of heads, $\rmO^{(i)}\in\sR^{1\times hd'}$,  $\omega$ is an activation function, 
$\theta^{(i)}\in\sR^{1\times d'}$, $\theta = ||_{i=1}^h \theta^{(i)}$,
$\rmK^{(i)}\in\sR^{c_*\times d'}$ and $\rmV^{(i)}\in\sR^{c_*\times d'}$ are the key and value vectors, respectively. 
$d'$ is the hidden dimension of $\rmV^{(i)}$, and $c_*$ is either $c_e$ for $\rmS_e^{\scriptscriptstyle{(\textbf{B})}}$ or $c_v$ for $\rmS_v^{\scriptscriptstyle{(\textbf{R})}}$. 
We employ two MLPs over $\rmS$ to obtain the key and value vectors, i.e., $\rmK^{(i)} = \text{MLP}_K^{(i)}(\rmS)$ and $\rmV^{(i)} = \text{MLP}_V^{(i)}(\rmS)$.

\subsubsection{Optimization for BHyGNN} 
A BHyGNN model contains multiple hypergraph convolutions and is optimized in an end-to-end manner.
The objective function $\gL_\text{sl}$ for supervised learning contains two components: the downstream task loss $\gL_\text{ce}$ and the variational loss $\gL_\text{var}$ for learning propagation actions: 
\begin{align}\label{eq: bhygnn loss}
    \gL_\text{sl} = \gL_\text{ce} + \alpha \sum_{l=1}^L \gL_\text{var}^{\scriptscriptstyle{(l)}},
\end{align}
where $\gL_\text{var}^{\scriptscriptstyle{(l)}}$ is the variational loss at the $l$-th BHyGNNConv layer, $L$ is the number of BHyGNNConv layers, and $\alpha$ is a hyperparameter to balance two losses.
Note that the variational loss $\gL_\text{var}$ encompasses the variational lower bound loss $\gL_\text{vlb}^{\scriptscriptstyle{\left(*\right )}}$ and regularization terms $\gL_\text{reg}^{\scriptscriptstyle{(\cdot)}}$ for both broadcast and receive actions, defined as:
\begin{align}
    & \gL_\text{var} = \gL_\text{vlb}^{\scriptscriptstyle{\left(\textbf{B}\right )}} + 
    \gL_\text{reg}^{\scriptscriptstyle{\left(\textbf{B}\right )}} + 
    \gL_\text{vlb}^{\scriptscriptstyle{\left(\textbf{R}\right )}} + 
    \gL_\text{reg}^{\scriptscriptstyle{\left(\textbf{R}\right )}}, \\
    &\gL_\text{vlb}^{\scriptscriptstyle{\left(*\right )}}  = \mathbb{E}_{q(\rmH_\gV |\gH )} \mathbb{E}_{q(\rmH_\gE |\gH )}[\log P(*| \rmH_\gV, \rmH_\gE)] \\ 
   & - \KL[q(\rmH_\gV | \gH ) | p(\rmH_\gV)]  - \KL[q(\rmH_\gE |\gH) | p(\rmH_\gE)], \\
   & \gL_\text{reg}^{\scriptscriptstyle{(*)}} = ||\lambda\rmM -\hat\rmM^{\scriptscriptstyle{(*)}}||,
\end{align}
where  $P(*| \rmH_\gV, \rmH_\gE)$ denotes the probability of either broadcast actions 
$P(\gB | \rmH_\gV, \rmH_\gE)$ or receive actions $P(\gR| \rmH_\gV, \rmH_\gE)$, with respect to $\gL_\text{vlb}^{\scriptscriptstyle{\left(\textbf{B}\right )}}$ or $\gL_\text{vlb}^{\scriptscriptstyle{\left(\textbf{R}\right )}}$, and $\KL[q(\cdot) | p(\cdot)]$ is the Kullback-Leibler divergence between variational posterior $q(\cdot)$ and prior $p(\cdot)$, with $p(\cdot)\sim\gN(0, I)$ as the standard Gaussian prior\footnote{Superscripts indicating specific actions, i.e., $\textbf{B}$ and $\textbf{R}$ for $\rmH_\gV$ and $\rmH_\gE$, respectively, are omitted for clarity.}.

According to Equation~\ref{eq: bhygnn loss}, BHyGNN leverages the training signals from downstream tasks to guide the learning of propagation actions, while these signals may be hard to obtain in various real-world scenarios.
This observation motivates us to introduce BHyGNN+ in the following section to tackle the aforementioned issue.
\subsection{BHyGNN+ for Self-Supervised Learning}
To address the limitation of BHyGNN, which relies on labeled data for learning propagation actions, we propose \modelname, a contrastive self-supervised learning (SSL) framework that models heterophilic hypergraphs without requiring downstream task labels.
The overall framework of BHyGNN+ is provided in Figure~\ref{fig: framework}.

\subsubsection{Hypergraph Augmentations}\label{sec: augmentation}
Inspired by the success of HyGCL methods~\cite{lee2023m, HyperGCL}, we incorporate hypergraph augmentations into \modelname to facilitate the SSL process.
We explore the following four classic augmentation techniques, denoted as $\gT$, to generate different views of hypergraphs. 

\textbf{Node Attribute Masking.} 
The underlying assumption is that corrupting attributes of a subset of nodes does not significantly affect the overall semantics.
To achieve this, we randomly select a fraction $p_\text{mask}$ of nodes $\gV_\text{mask}$ and mask their features, resulting in the augmented feature set $\widetilde{\gX_\gV}$.
Formally, we have:
\begin{align}
    \widetilde{\gX_\gV} = \{ \rmX_v + \epsilon_v : v \in \gV_\text{mask}, \epsilon_v \sim \gN(0, I) \}  \cup \{\rmX_u : \rmX_u \in \gV \setminus \gV_\text{mask}\}.
\end{align}

\textbf{Hyperedge Perturbation.}
Merely masking node attributes may not be sufficient to capture the complex relationships in heterophilic hypergraphs. 
To this end, we propose to perturb hyperedges to augment the hypergraphs.
Note that, unlike existing works that also allow adding nodes to hyperedges, we only remove nodes from hyperedges to avoid introducing noisy connections in heterophilic hypergraphs.
Specifically, we obtain the augmented hyperedges $\widetilde{\gE}$ by randomly selecting a fraction $p_\text{pert}$ of hyperedges $\gE_\text{pert}$, and for each hyperedge $e \in \gE_\text{pert}$, we randomly remove a subset of nodes $\gV_\text{perb}^{(e)}$.
The mathematical formulation is as follows:
\begin{align}
    \widetilde{\gE} = \{e \setminus \gV^{(e)}_\text{perb} : e \in \gE_\text{pert},  \gV^{(e)}_\text{perb} \subset e \} \cup \{ e: e \in \gE \setminus \gE_\text{pert}\}.
\end{align}

\textbf{Hyperedge Dropping.} 
In this augmentation, a fraction \(p_\text{drop}\) of hyperedges $\gE_\text{drop}$ are randomly removed from the hypergraph $\gH$.
The underlying idea is that removing some structural information would not significantly affect the overall semantics of the hypergraph.
Mathematically, we have:
\begin{align}
    \widetilde{\gE} = \{\gE \setminus \gE_\text{drop}\}.
\end{align}

\textbf{Node Dropping.} We also consider node dropping as discussed in GCL~\cite{you2020graph}. 
Specifically, we randomly remove a fraction $p_\text{drop}$ of nodes $\gV_\text{drop}$ from the hypergraph $\gH$, along with all their incidences in the hyperedges $\gE$.
\begin{align}
    \widetilde{\gV} = \{\gV \setminus \gV_\text{drop}\}, \widetilde{\gE} = \{ e \setminus \gV_\text{drop} : e \in \gE\}.
\end{align}

\subsubsection{Self-Supervised Learning over Hypergraph Duals}
The core idea of \modelname is to leverage hypergraph duality (see Definition~\ref{def: dual-hypergraph}) to design an SSL framework that captures structural information from both the original and dual perspectives.

Given a hypergraph $\gH$, following existing methods~\cite{HyperGCL, kim2024hypeboy}, \modelname first augments both the hypergraph $\gH$ and its dual $\gH^*$ into multiple views.
Mathematically, we have:
\begin{align}
    \tilde\gH_i= \text{Aug}_i(\gH), \, \tilde\gH_i^* = \text{Aug}_i(\gH^*),  \text{Aug}_i(\cdot) \in \gT.
\end{align}
Here, $\gT$ is a set of hypergraph augmentation techniques, which we discuss in Section~\ref{sec: augmentation}.

These augmented hypergraphs are fed into the HyGNN encoder $f(\cdot)$ to learn hypergraph representations. 
Unlike existing HyGCL methods that perform contrastive learning at the node level, \modelname directly contrasts global hypergraph representations. 
Specifically, the global representation of each hypergraph $\rmZ_\gH$ is obtained via a $\text{Readout}(\cdot)$ function, which is computed as: 
\begin{align}
    \rmZ_{\widetilde{\gH_i}} = \text{Readout}\left(f(\widetilde{\gH_i})\right),
    \rmZ_{\widetilde{\gH_i}^*} =  \text{Readout}\left(f(\widetilde{\gH_i}^*)\right).
\end{align}
Note that the global representations $\rmZ_{\gH^*}$ for all augmented hypergraphs $\widetilde{\gH}_*$ are obtained in the same way, and these representations are merely used to guide model optimization. 
At inference time, we directly feed the original hypergraph into the trained HyGNNs $f(\cdot)$ to obtain the hypergraph representations, and these representations are further passed into a classifier for downstream tasks.

\subsubsection{Optimization for BHyGNN+}
\modelname employs a contrastive loss $\gL_\text{con}$ to guide the learning process.
Specifically, the objective is to maximize the similarity between the augmented dual hypergraph representations $\rmZ_{\widetilde{\gH_i}^*}$ and the corresponding augmented hypergraph representations $\rmZ_{\widetilde{\gH_i}}$ using cosine similarity:
\begin{align}
    \gL_\text{con} = \frac{1}{D} \sum_{i=0}^D \frac{\rmZ_{\widetilde{\gH_i}^*} \rmZ_{\widetilde{\gH_i}}^T}{||\rmZ_{\widetilde{\gH_i}^*}||||\rmZ_{\widetilde{\gH_i}}||}.
\end{align}

Therefore, the final objective function for BHyGNN+ simply replaces the downstream task loss $\gL_\text{ce}$ in Equation~\ref{eq: bhygnn loss} with the contrastive loss $\gL_\text{con}$:
\begin{align}\label{eq: bhygnnplus_loss}
    \gL_\text{ssl} = \gL_\text{con} + \alpha \sum_{l=1}^L \gL_\text{var}^{\scriptscriptstyle{(l)}},
\end{align}
where $\gL_\text{var}^{\scriptscriptstyle{(l)}}$ is the same variational loss as in BHyGNN, and $\alpha$ is a hyperparameter.

By optimizing this objective, \modelname learns to capture hypergraph structure and produce high-quality node and hyperedge representations without relying on external labels. These learned representations can then be used for various downstream tasks such as node classification, clustering, or link prediction.

\section{Experiments}\label{sec: experiment}
To comprehensively evaluate the performance of \modelname, we conduct extensive experiments on eleven datasets with thirteen baseline methods. 
In this section, we first discuss our experimental setup, including the employed benchmark datasets, baseline methods, and experiment settings.
Afterward, we discuss the main experimental comparison with baseline methods across heterophilic and homophilic benchmark hypergraphs under various settings, as well as a hyperparameter sensitivity analysis.
\subsection{Experimental Setup}
\subsubsection{Benchmark Datasets}
We employ one synthetic dataset and ten benchmark hypergraph datasets: Senate~\cite{chodrow2021hypergraph}, Congress~\cite{Benson-2018-simplicial, fowler2006legislative, Fowler-2006-connecting}, House~\cite{chodrow2021hypergraph}, Walmart~\cite{amburg2020clustering}, Co-authorship networks (Cora-CA and DBLP), Co-citation networks (Cora, Pubmed, Citeseer)~\cite{HyperGCN}, and Twitter~\cite{ma2023hypergraph}. 
These benchmark datasets cover a wide range of domains and can be categorized into heterophilic and homophilic hypergraphs based on their homophily scores.

\textbf{Heterophilic Hypergraphs.} 
Senate, Congress, House, and Walmart are heterophilic hypergraph datasets since their average homophily scores are less than 0.8. 
For Senate, Congress, and House datasets, each node is labeled with political party affiliation. In the Walmart network,  nodes represent products that are purchased at Walmart, and hyperedges represent sets of products purchased together. 
All four benchmark hypergraph datasets mentioned above do not contain node features. 
We employ the same setting as existing works~\cite{wang2022equivariant, veldt2021higher}, which introduce node features through a label-dependent Gaussian distribution~\cite{deshpande2018contextual}.
Specifically, the node feature dimension is fixed to 100, and each feature vector is the one-hot encoding of the node label with added Gaussian noise $\mathcal{N}(0, \sigma^2)$, where the noise standard deviation $\sigma$ is set to $0.6$ or $1.0$.

\textbf{Homophilic Hypergraphs.}
The rest of the hypergraph datasets, i.e., Co-authorship networks, Co-citation networks, and Twitter, are homophilic datasets. 
For Co-authorship benchmark hypergraph datasets, documents co-authored by an author are connected by a hyperedge. For Co-citation benchmark hypergraph datasets, all documents referenced by a document are connected by a hyperedge. In all five citation hypergraphs, node features are the bag-of-words representations of the corresponding documents, and node labels are the paper classes.
Twitter is derived from the benchmark dataset Twitter-HyDrug~\cite{ma2023hypergraph}, where nodes are Twitter users related to illicit drug trafficking, and each hyperedge represents an explicit or implicit relationship among a set of Twitter users. 
We use the identical node attribute features and hypergraph structure in Twitter-HyDrug. The task for the hypergraph dataset Twitter is drug role classification, i.e., the node labels are drug seller, drug buyer, drug user, and drug discussant. 

\textbf{Synthetic Heterophilic Hypergraphs.}
Besides the existing benchmark hypergraph datasets, to further evaluate our model on heterophilic hypergraphs, similar to ED-HNN~\cite{wang2022equivariant}, we leverage the contextual hypergraph stochastic block model~\cite{deshpande2018contextual, NIPS2014_8f121ce0} to construct a synthetic heterophilic hypergraph. 
To construct a more challenging synthetic dataset, unlike ED-HNN, which merely introduces two classes, we draw four classes of 500 nodes each and randomly sample 250 hyperedges for each node class (four majority classes and 1,000 hyperedges in total). Each hyperedge contains 15 nodes, five nodes from the majority class and ten nodes from the rest of the classes (minority classes). 

\subsubsection{Baseline Methods}
We adopt MLP, supervised HyGNNs (i.e., HGNN~\cite{HGNN}, HyperGCN~\cite{HyperGCN}, HNHN~\cite{HNHN}, HCHA~\cite{HCHA}, UniGCNII~\cite{UniGNN}, AllSet~\cite{Allset}, ED-HNN~\cite{wang2022equivariant}, SheafHGNN~\cite{duta2024sheaf}), and self-supervised HyGNNs (i.e., HyperGCL~\cite{HyperGCL}, TriCL~\cite{lee2023m}, HyGCL-ADT~\cite{qian2024dual}, and HypeBoy~\cite{kim2024hypeboy}) as our baseline methods.

\subsubsection{Experiment Settings}
We use accuracy as the evaluation metric for both our model and baseline methods. 
To conduct experiments in challenging and practical settings, we split the data into training/validation/test samples using 20\%/20\%/60\%. 
Moreover, we run each method ten times with 500 epochs and report the average score with standard deviation (std). 
All experiments are conducted under the environment of Ubuntu
22.04.3 OS plus an Intel i9-12900K CPU, four Nvidia A40 graphics cards, and 48 GB of RAM.
We utilize Adam as the optimizer and run a grid search on hyperparameters for each model to obtain the best performance.
We define the specific ranges to find the optimal hyperparameters for every model. For instance, the range of hidden dimensions for layers is $\{64, 128, 256, 512, 1024\}$, the range of weight decays is $\{0.0, 0.01, 0.001\}$, and the range of learning rate is $\{0.1, 0.01, 0.001, 0.0001\}$. 

For a fair comparison, we follow existing works~\cite{qian2024dual, kim2024hypeboy} to conduct experiments over supervised and unsupervised methods.
Specifically, supervised methods are trained end-to-end with an appended two-layer MLP as the classifier. 
In contrast, unsupervised methods are first pre-trained to learn hypergraph representations. 
These representations then serve as input to a separate two-layer MLP classifier for the downstream task. 
To ensure the evaluation solely assesses the quality of the learned representations, the parameters of the pre-trained unsupervised models are frozen during the classifier training phase.

\begin{table*}[t]
    \small
    \caption{Performance comparison (Mean accuracy \%) of baseline methods for node classification on heterophilic hypergraph datasets. 
    OOM indicates an out-of-memory error.
    For each group, the bolded numbers indicate the best result and underlined numbers represent the runner-up performance within each group. 
    % The full results with std. are provided in Appendix~\ref{tab: full performance het}.
    } 
    \label{tab: performance het}
    \centering
    \resizebox{\linewidth}{!}{
    \begin{tabular}{llcccccccccc}
         \toprule
          & & \multicolumn{2}{c}{\textsc{Senate}} & \multicolumn{2}{c}{\textsc{Synthetic}} & \multicolumn{2}{c}{\textsc{Congress}} & \multicolumn{2}{c}{\textsc{House}} & \multicolumn{2}{c}{\textsc{Walmart}} \\
          \cmidrule(lr){1-2}\cmidrule(lr){3-4}\cmidrule(lr){5-6}\cmidrule(lr){7-8}\cmidrule(lr){9-10}\cmidrule(lr){11-12}
          \multicolumn{2}{l}{Num. of Nodes} & 282 & 282 & 2,000 & 2,000 & 1,718 & 1,718 &1,290 & 1,290 &  88,860 & 88,860\\
         \multicolumn{2}{l}{Num. of Edges} & 315 & 315 & 1,000 & 1,000 & 83,105 & 83,105 & 340 & 340 & 69,906 & 69,906\\ 
         \multicolumn{2}{l}{Average $h(v)$}  & 0.50 & 0.50  & 0.28 & 0.28 & 0.55 & 0.55 & 0.51 & 0.51 & 0.53 & 0.53 \\ 
         \multicolumn{2}{l}{Average $h(e)$} & 0.55 & 0.55 & 0.33 & 0.33 & 0.79 & 0.79 & 0.58 & 0.58 &  0.75 & 0.75 \\
         \multicolumn{2}{l}{Noise std. $\sigma$} & 0.6 & 1.0 & 0.6 & 1.0 & 0.6 & 1.0 & 0.6 & 1.0 & 0.6 & 1.0 \\
          \cmidrule(lr){1-2}\cmidrule(lr){3-4}\cmidrule(lr){5-6}\cmidrule(lr){7-8}\cmidrule(lr){9-10}\cmidrule(lr){11-12}
         \multirow{10}{*}{\rotatebox{90}{Supervised}} & MLP & 62.24 & 50.35 & 50.00 & 37.48 & 79.45 & 65.91 & 77.12 & 64.07 & 62.23 & 44.69 \\
          & HGNN & 60.06& 49.35 & 42.42& 37.90 & 90.91& 88.08 & 61.24& 57.24 & 77.19& 61.34 \\
         & HyperGCN & 55.00& 51.82 & 41.61& 32.51 & 84.81& 83.32 & 75.62& 62.43 & 62.02& 49.61 \\
         & HNHN& 62.18& 54.71 & 49.67& 37.11 & 89.71& 82.97 & 68.36& 65.16 & 68.68& 58.01 \\
         & HCHA & 47.71& 46.20 & 32.50& 27.18 & 91.04& 89.81 & 61.28& 56.98 & 76.55& 61.83 \\
         & UniGCNII & 60.06& 52.24 & 49.62& 37.13 & 92.91& 89.56 & 78.64& 65.45 & 72.36& 63.72 \\
         & AllSet & 65.47& 51.76 & 52.84& 42.78 & 93.65& 88.65 & 78.81& 65.20 & \underline{78.74}& \underline{65.35} \\
         & ED-HNN & \underline{65.53}& \underline{55.47} & \underline{55.96}& 43.59 & \underline{94.20}& \underline{92.07} & 79.01& 65.70 & 78.15& 65.07 \\
         & SheafHGNN & 64.35& 54.32 & 55.42& \underline{43.97} & 90.72& 91.07 & \underline{79.75}& \underline{65.93} & OOM & OOM \\
         & \cellcolor{Gray} BHyGNN & \cellcolor{Gray} \textbf{67.87}& \cellcolor{Gray} \textbf{58.41} & \cellcolor{Gray} \textbf{58.10}& \cellcolor{Gray} \textbf{47.32} & \cellcolor{Gray} \textbf{95.45}& \cellcolor{Gray} \textbf{93.72} & \cellcolor{Gray} \textbf{80.23}& \cellcolor{Gray} \textbf{67.36} & \cellcolor{Gray} \textbf{79.94}& \cellcolor{Gray} \textbf{66.85} \\
          \cmidrule(lr){1-2}\cmidrule(lr){3-4}\cmidrule(lr){5-6}\cmidrule(lr){7-8}\cmidrule(lr){9-10}\cmidrule(lr){11-12}
        \multirow{5}{*}{\rotatebox{90}{Unsupervised}} & HyperGCL & \underline{66.89} & \underline{56.28} & 56.73 & 44.07 & 94.84 & 92.73 & 79.67 & \underline{66.95} & \underline{80.83} & \underline{67.28} \\ 
         % & HyperGRL \\ 
         & TriCL & 64.50 & 53.38 & \underline{57.23} & \underline{45.83} & \underline{95.18} & \underline{93.73} & \underline{80.55} & 65.81 & 79.27 & 66.38 \\
        & HyGCL-ADT & 64.85& 52.93 & 55.92& 42.31 & 93.10& 91.86 & 79.43& 64.72 & 78.41& 65.34 \\
         & HypeBoy & 63.47& 53.17 & 53.74& 42.26 & 92.34& 90.52 & 79.32& 65.35 & 76.42& 64.28 \\
          % \cmidrule(lr){1-2}\cmidrule(lr){3-4}\cmidrule(lr){5-6}\cmidrule(lr){7-8}\cmidrule(lr){9-10}\cmidrule(lr){11-12}
       & \cellcolor{Gray} BHyGNN+ & \cellcolor{Gray} \textbf{68.13} & \cellcolor{Gray} \textbf{60.38} & \cellcolor{Gray} \textbf{58.97} & \cellcolor{Gray} \textbf{48.27} & \cellcolor{Gray} \textbf{96.33} & \cellcolor{Gray} \textbf{95.25} & \cellcolor{Gray} \textbf{80.93} & \cellcolor{Gray} \textbf{69.17} & \cellcolor{Gray} \textbf{80.65} & \cellcolor{Gray} \textbf{68.51} \\
         \bottomrule
    \end{tabular}
    }
\end{table*}

\subsection{Results and Analysis}
\subsubsection{Performance over Heterophilic Hypergraphs}
Table~\ref{tab: performance het} presents the performance comparison of \modelname with baseline methods on five heterophilic hypergraph datasets.
According to the table, we find that: 
(i) MLP that ignores hypergraph structures can outperform several HyGNNs
in heterophilic hypergraphs. This finding implies that the heterophilic hypergraph structures may negatively affect the
performance of HyGNNs.
(ii) Most unsupervised learning methods outperform all supervised methods. This indicates the necessity to adopt SSL over heterophilic hypergraphs. 
(iii) \modelname outperforms all baseline methods, including BHyGNN, demonstrating its effectiveness on heterophilic hypergraphs.

\begin{table*}[t]
    \small
    \caption{Performance comparison (Mean accuracy \%) of baseline methods for node classification on homophilic hypergraph datasets. 
    For each group, bolded numbers indicate the best result, and underlined numbers represent the runner-up performance within each group.}
    \label{tab: performance homo}
    \centering
    \resizebox{\linewidth}{!}{
    \begin{tabular}{llcccccccccccc}
        \toprule
         & & \multicolumn{2}{c}{\textsc{Twitter}} & \multicolumn{2}{c}{\textsc{Citeseer}}& \multicolumn{2}{c}{\textsc{DBLP}}& \multicolumn{2}{c}{\textsc{Cora}} & \multicolumn{2}{c}{\textsc{Cora-CA}} & \multicolumn{2}{c}{\textsc{Pubmed}} \\
        \cmidrule(lr){1-2}\cmidrule(lr){3-4}\cmidrule(lr){5-6}\cmidrule(lr){7-8}\cmidrule(lr){9-10}\cmidrule(lr){11-12}\cmidrule(lr){13-14}
        \multicolumn{2}{l}{Num. of Nodes}  & 2,936 & 2,936 & 3,312 &  3,312 & 41,302 & 41,302 & 2,708 & 2,708 & 2,708 & 2,708 & 19,717 & 19,717  \\ 
        \multicolumn{2}{l}{Num. of Edges}  & 35,502 & 35,502 & 1,070 & 1,079 &  22,363 & 22,363 &1,072 & 1,072 & 1,579 & 1,579 & 7,963 & 7,963   \\ 
        \multicolumn{2}{l}{Average $h(v)$} & 0.41 & 0.82  & 0.42 & 0.83 & 0.49 & 0.87 & 0.40 & 0.80 & 0.45 & 0.90 & 0.48 & 0.95   \\ 
        \multicolumn{2}{l}{Average $h(e)$} & 0.45 & 0.90 & 0.42 & 0.83 & 0.47 & 0.93  & 0.44 & 0.88 & 0.43 & 0.86 & 0.44 & 0.88   \\ 
        \multicolumn{2}{l}{Hypergraph Type} & \textsc{Het.} & \textsc{Homo.}  & \textsc{Het.} & \textsc{Homo.}  & \textsc{Het.} & \textsc{Homo.}  & \textsc{Het.} & \textsc{Homo.}  & \textsc{Het.} & \textsc{Homo.}  & \textsc{Het.} & \textsc{Homo.}\\ 
        \cmidrule(lr){1-2}\cmidrule(lr){3-4}\cmidrule(lr){5-6}\cmidrule(lr){7-8}\cmidrule(lr){9-10}\cmidrule(lr){11-12}\cmidrule(lr){13-14}
        \multirow{10}{*}{\rotatebox{90}{Supervised}} & MLP & 67.38 & 67.38 & \underline{68.05} & 68.05 & 83.76 & 83.76 & \underline{68.45} & 68.45 & 69.45 & 69.45 & 84.45 & 84.45 \\
        & HGNN& 67.17 & 68.52 & 55.10 & 69.32 & 68.53 & 90.38 & 58.64 & 76.11 & 62.94 & 79.42 & 80.91 & 85.82 \\
        & HyperGCN& 55.68 & 69.29 & 56.87 & 69.13 & 65.80 & 88.37 & 56.41 & 73.89 & 56.59 & 75.12 & 64.96 & 86.31 \\
        & HNHN & 62.08 & 67.78 & 66.65 & 68.36 & 81.76 & 86.42 & 62.79 & 71.52 & 64.69 & 72.12 & 83.62 & 85.92 \\
        & HCHA  & 67.63 & 72.04 & 53.33 & 68.84 & 67.91 & 90.27 & 55.71 & 75.97 & 63.27 & 79.23 & 75.87 & 83.53 \\
        & UniGCNII & 67.66 & 69.57  & 60.38 & 70.21 & 81.80 & 90.53 & 58.08 & \textbf{76.25} & 61.66 & 78.20 & 85.24 & 86.31 \\
        & AllSet  & \underline{69.16} & 70.64 & 67.91 & 69.71 & \underline{84.06} & 90.69 & 67.54 & 74.26 & \underline{70.04} & 78.30 & 85.04 & 86.38 \\
        & ED-HNN & 69.07 & \underline{73.47}  & 66.30 & 69.82 & 83.90 & \underline{90.85} & 67.13 & 75.78 & 69.90 & 79.88 & \underline{86.04} & \underline{87.26} \\
        & SheafHGNN & 68.95 & 72.87  & 66.16 & \textbf{70.57} & 82.17 & 90.29 & 67.42 & \underline{76.19} & 69.82 & \textbf{81.14} & 84.16 & 85.41\\ 
        & \cellcolor{Gray} BHyGNN & \cellcolor{Gray} \textbf{72.86} & \cellcolor{Gray} \textbf{75.61}  & \cellcolor{Gray} \textbf{68.95} & \cellcolor{Gray} \underline{70.12} & \cellcolor{Gray} \textbf{85.05} & \cellcolor{Gray} \textbf{90.93} & \cellcolor{Gray} \textbf{69.26} & \cellcolor{Gray} 75.87 & \cellcolor{Gray} \textbf{70.99} & \cellcolor{Gray} \underline{80.72}  & \cellcolor{Gray} \textbf{86.90} & \cellcolor{Gray} \textbf{88.02} \\
        \cmidrule(lr){1-2}\cmidrule(lr){3-4}\cmidrule(lr){5-6}\cmidrule(lr){7-8}\cmidrule(lr){9-10}\cmidrule(lr){11-12}\cmidrule(lr){13-14}
        \multirow{5}{*}{\rotatebox{90}{Unsupervised}} 
        & HyperGCL & {69.97} & 73.13 & {68.34} & 70.21 & \underline{85.10} & 90.83 & {69.65} & 75.27 & \underline{70.29} & 79.39 & \underline{86.80} & 86.98 \\ 
        % & HyperGRL \\ 
        & TriCL & \underline{72.97} & \underline{73.18} & \underline{68.94} & \underline{72.18} & 83.37 & {91.12} & \underline{69.84} & 76.21 & 69.19 & {79.92} & 85.28 &  {87.04}  \\ 
        & HyGCL-ADT & {70.08}  & 72.46 & 66.51 & 71.03 & 83.76 & 90.56 & 67.01 & 76.35 & 69.72 & \textbf{80.52} & 85.73 & 86.89 \\ 
        & HypeBoy  & 68.77 & 71.36 & 67.05 & 70.98 & 81.72 & \textbf{91.26} & 66.82 & \underline{76.40} & 68.39 & 80.04 & 85.29 & \underline{87.94} \\
        % \cmidrule(lr){1-2}\cmidrule(lr){3-4}\cmidrule(lr){5-6}\cmidrule(lr){7-8}\cmidrule(lr){9-10}\cmidrule(lr){11-12}\cmidrule(lr){13-14} 
        & \cellcolor{Gray} BHyGNN+ & \cellcolor{Gray} \textbf{73.17} & \cellcolor{Gray} \textbf{75.92} & \cellcolor{Gray} \textbf{69.79} & \cellcolor{Gray} \textbf{70.64}  & \cellcolor{Gray} \textbf{86.17} & \cellcolor{Gray} \underline{91.17} & \cellcolor{Gray} \textbf{70.92} & \cellcolor{Gray} \textbf{76.42} & \cellcolor{Gray} \textbf{71.84} & \cellcolor{Gray} \underline{80.31} & \cellcolor{Gray} \textbf{87.12} & \cellcolor{Gray} \textbf{88.47}  \\
        \bottomrule
    \end{tabular}
    }
\end{table*}

\subsubsection{Performance over Homophilic Hypergraphs}
To further evaluate the effectiveness of our model, we compare baseline models with \modelname on benchmark homophilic hypergraph datasets.
Moreover, inspired by hypergraph augmentation methods~\cite{HyperGCL}, we augment the benchmark homophilic hypergraph datasets to simulate heterophily and report the performance of baseline methods and our models.
Table~\ref{tab: performance homo} lists the performance of baseline methods and \modelname under the aforementioned two settings, i.e., Het. and Homo.
Based on the result, we make the following conclusions: 
(i) The feature-based method MLP performs better than at least three HyGNNs in each augmented dataset. Moreover, MLP achieves the best performance among all baseline methods on the augmented Citeseer and Cora datasets. These findings again verify that heterophilic hypergraph structures can negatively affect HyGNNs.
(ii) Again, unsupervised methods outperform supervised baseline methods to some extent, indicating the contribution of contrastive learning toward hypergraph representation learning. 
(iii) \modelname demonstrates better performance than baseline methods on heterophilic hypergraphs and most homophilic datasets, validating its superiority in both settings. 
\subsubsection{Additional Experiments}
Following existing studies~\cite{Allset, duta2024sheaf, HyperGCL}, we conduct additional experiments over two settings, i.e., $50\%/25\%/25\%$ and $10\%/10\%/80\%$
for training, validation, and testing, respectively, and the result is provided in Table~\ref{tab: additional exp}.
According to the table, we find that:
(i) With a high proportion of training samples, i.e., $50\%/25\%/25\%$, the performance gap between self-supervised and supervised baselines (e.g., HyperGCL, HyGCL-ADT, and HypeBoy) is marginal. 
This may be due to the training data already being sufficient for supervised HyGNNs to learn informative representations within hypergraphs. 
(ii) In a low training sample setting, \modelname exhibits a significant improvement over BHyGNN, likely because BHyGNN struggles to learn effective propagation actions under limited supervision.

\begin{table}[t]
    \caption{Performance comparison in additional split settings.}   \label{tab: additional exp}
    \vspace{-1mm}
    \centering
    \begin{minipage}[t]{0.5\linewidth}
         \subcaption{Experimental result with split 50\%/25\%/25\%.}
        \resizebox{\linewidth}{!}{
        \begin{tabular}{lcccc}
        \toprule
        \multirow{2}{*}{Model} & \textsc{Congress} & \textsc{Walmart} & \textsc{Twitter} & Cora\\ 
        & (1.0)  & (1.0) & \textsc{(Homo.)} & \textsc{(Homo.)}\\ 
        \midrule
        AllSet &  92.16 & 65.46 & 73.62 & 78.58\\ 
        ED-HNN  &  \underline{95.00} & 66.91 & 75.10 & 80.31 \\ 
        % SheafHGNN & 91.81 & OOM & 74.21 & \textbf{81.30}\\ 
        HyperGCL  & 94.73 & 67.12 & \underline{75.73} &  80.26 \\ 
        HyGCL-ADT & 93.83 & 66.48 & 74.02 & 79.28\\ 
        HyperBoy & 93.91 & \underline{67.94} & 73.54 & \textbf{82.39} \\
        \midrule
        \rowcolor{Gray} BHyGNN & 96.41 & 68.82 & 77.76 & 81.02\\ 
        \rowcolor{Gray} BHyGNN+ & \textbf{96.83} & \textbf{69.10} & \textbf{78.12} & \underline{81.16} \\
          \bottomrule
        \end{tabular}
        }
   \end{minipage}%
    \begin{minipage}[t]{0.5\linewidth}
        \subcaption{Experimental result with split 10\%/10\%80/\%.}
        \resizebox{\linewidth}{!}{
        \begin{tabular}{lcccc}
        \toprule 
        \multirow{2}{*}{Model} & \textsc{Congress} & \textsc{Walmart} & \textsc{Twitter} & Cora\\ 
        & (1.0)  & (1.0) & \textsc{(Homo.)} & \textsc{(Homo.)}\\ 
        \midrule
        AllSet & 87.91 & 59.13  & 69.57 & 67.93\\ 
        ED-HNN & 89.17 & 60.23 & 72.20 & \underline{74.31} \\ 
        % SheafHGNN & & & &\\
        HyperGCL & \underline{90.20} & \underline{60.46}  & \underline{73.07} & 73.12\\
        HyGCL-ADT & 89.37 & 59.96 & 72.19 & 72.37 \\
        HypeBoy & 88.39 & 60.45 & 70.60 & 74.21\\
        \midrule
        \rowcolor{Gray} BHyGNN & 90.23 & 60.92 & 73.18  & 70.82 \\
        \rowcolor{Gray} BHyGNN+ & \textbf{93.16} & \textbf{63.89}  & \textbf{74.28} & \textbf{74.92} \\
        \bottomrule
        \end{tabular}
        }
    \end{minipage}
\end{table}

\subsubsection{Hyperparameter Sensitivity Analysis}
\begin{figure}
    \centering
    \includegraphics[width=0.9\linewidth]{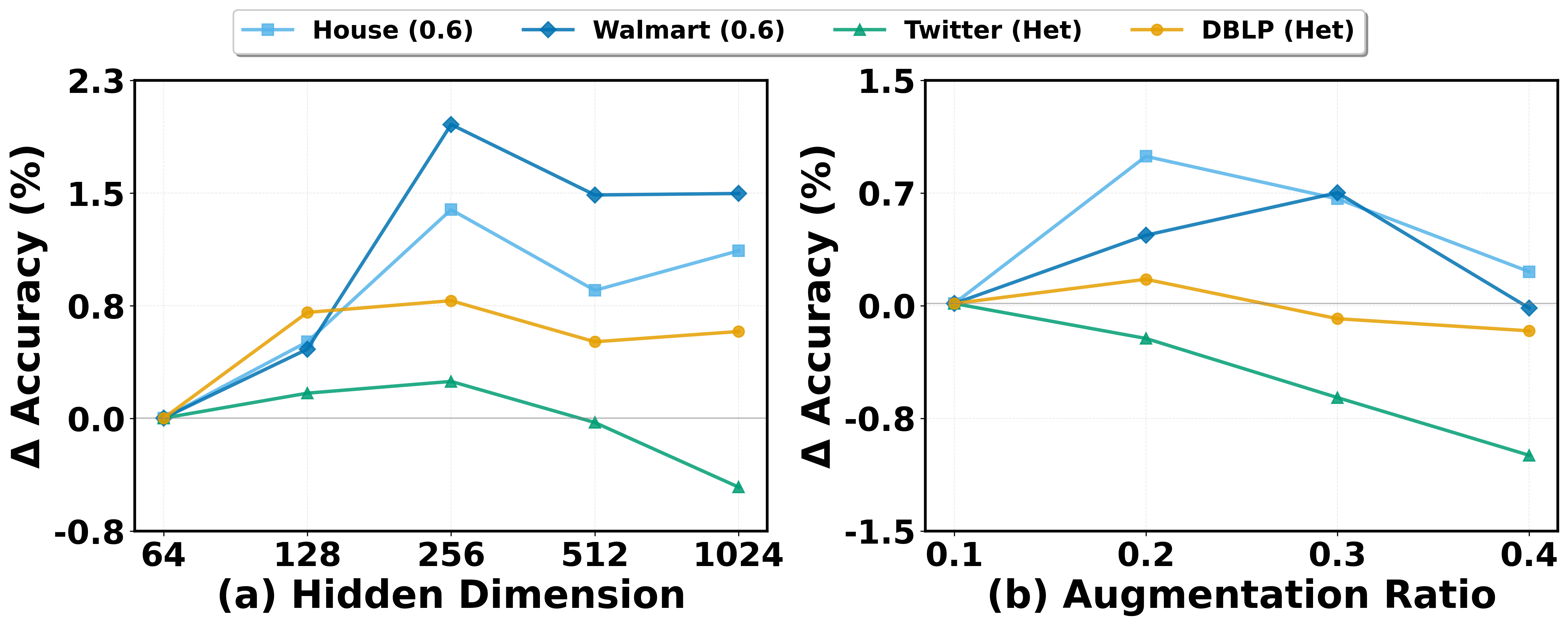}
    \caption{Hyperparameter sensitivity analysis on (a) hidden dimensions and (b) augmentation ratio across various benchmark datasets.}
    \label{fig: sensitive}
\end{figure}

We also conduct a sensitivity analysis among hyperparameters, i.e., hidden dimensions and the augmentation fraction $p$.
To clearly demonstrate the performance trend across different hyperparameters, we normalize the results based on the baseline configuration, i.e., hidden dimension $d=64$ and augmentation ratio $p=0.1$. 
The result is provided in Figure~\ref{fig: sensitive}.

\textbf{Hidden Dimensions.} 
We first conduct experiments with hyperparameter hidden dimension $d$ with range $\{64, 128, 256, 512, 1024\}$, and fix all other hyperparameters. 
According to Figure~\ref{fig: sensitive}(a), we find that as the hidden dimension increases, the model performance generally improves; however, further increasing the hidden dimension does not lead to continuous improvement.
Despite variations, our model maintains stable performance across different dimensions, demonstrating the robustness of \modelname on both heterophilic and homophilic hypergraphs.

\textbf{Augmentation Ratio.} 
Next, we report performance for augmentation ratios selected from \{0.1, 0.2, 0.3, 0.4\}.
Similarly, we fix all other hyperparameters and evaluate performance as a function of the augmentation ratio.
According to Figure~\ref{fig: sensitive}(b), \modelname achieves its best performance at an augmentation ratio $p = 0.2$ on most datasets.
Increasing $p$ beyond this point leads to a slight decline in performance.
This observation suggests that moderate augmentation introduces beneficial perturbations that enhance generalization, while excessive augmentation may inject too much noise, distorting the underlying signal and hindering learning.

\section{Conclusion}
In this work, we presented \modelname, a self-supervised learning framework for heterophilic hypergraph representation learning.
By leveraging \textbf{hypergraph duality}, i.e., nodes and hyperedges interchange roles, \modelname learns meaningful representations through contrasting augmented views of a hypergraph with its dual, eliminating the need for both ground-truth labels and negative samples.
Extensive experiments on eleven benchmark datasets demonstrate that \modelname consistently outperforms state-of-the-art supervised and unsupervised baselines across heterophilic and homophilic settings, validating the effectiveness of our approach.

\section{Limitations and Future Work}
While \modelname demonstrates strong performance, several avenues remain for future exploration.
First, the current framework focuses on node-level downstream tasks; extending it to hyperedge-level or hypergraph-level tasks could broaden its applicability.
Second, exploring augmentation strategies specifically tailored for heterophilic structures may further enhance performance.
Third, investigating the theoretical foundations underlying the effectiveness of hypergraph duality in heterophilic settings could yield deeper insights.

  \clearpage
  \bibliographystyle{plainnat}
  \bibliography{citations}

  % \clearpage
  % \appendix
  % \input{sections/appendix}

  %%%%%%%%%%%%%%%%%%%%%%%%%%%%%%%%%%%%%%%%%%%%%%%%%%%%%%%%%%%%
\end{document}